%% file: main.tex
\documentclass[lettersize,journal]{IEEEtran}
\usepackage{amsmath,amsfonts}
\usepackage{algorithmic}
\usepackage{algorithm}
\usepackage{array}
\usepackage[caption=false,font=normalsize,labelfont=sf,textfont=sf]{subfig}
\usepackage{textcomp}
\usepackage{stfloats}
\usepackage{url}
\usepackage{verbatim}
\usepackage{graphicx}
\usepackage{cite}
\usepackage{multirow}
\begin{document}

\title{EgoViT: Pyramid Video Transformer for Egocentric Action Recognition}

\author{Chenbin Pan, Senem Velipasalar,~\IEEEmembership{Senior Member,~IEEE \vspace{-0.4cm}}
\thanks{Authors are with the Dept. of Electrical Engineering and Computer Science, Syracuse University, NY. e-mail:\{cpan14,svelipas\}@syr.edu}%
\thanks{Manuscript received 2022; revised .}
}

\markboth{Journal of \LaTeX\ Class Files,~Vol., No., 2022}%
{Shell \MakeLowercase{\textit{et al.}}: A Sample Article Using IEEEtran.cls for IEEE Journals}


\maketitle

\begin{abstract}
\input{0_abstract}

\end{abstract}


\section{Introduction}
\input{1_introduction}

\section{Related Work}
\input{2_related_work}

\section{Proposed Model}
\input{3_model}

\section{Experiments}\label{sec:4_exp}
\input{4_experiments}

\section{Conclusion}

\input{5_conclusion}

\section*{Acknowledgments}
This work was supported in part by the National Science Foundation under Grant 1816732.



\bibliographystyle{IEEEtran}
\bibliography{egbib}

\newpage

 
\vspace{11pt}

\begin{IEEEbiography}[{\includegraphics[width=1in,height=1.25in,clip,keepaspectratio]{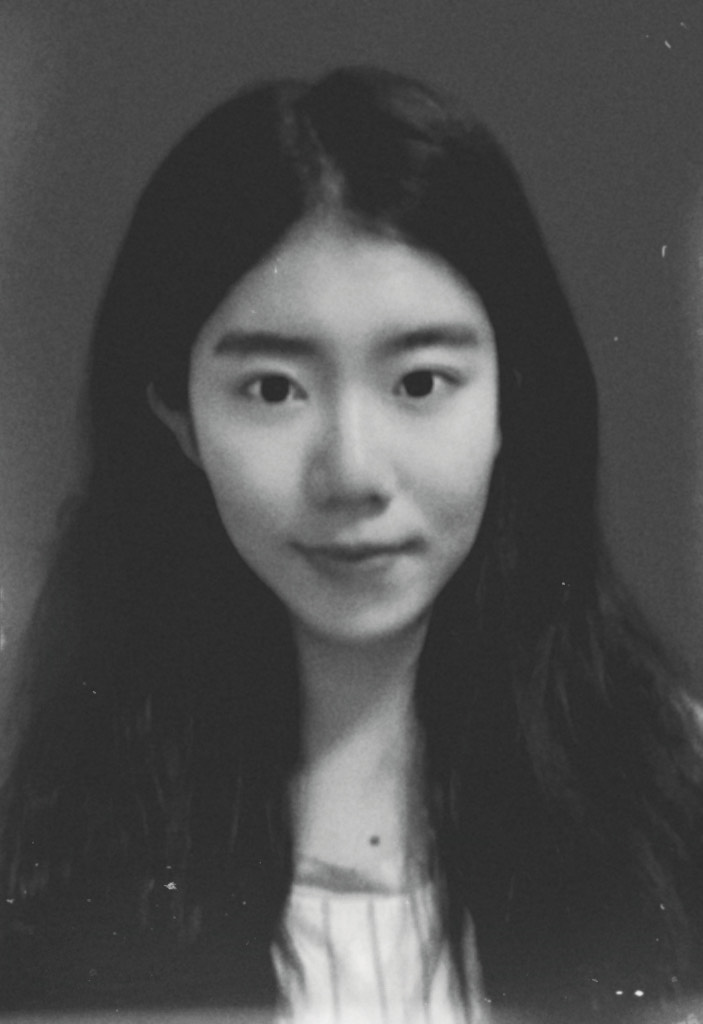}}]{Chenbin Pan} received the B.S. degree in electronic science and technology from the Beijing Institute of Technology, Beijing, China, in 2019. She is currently working towards the Ph.D. degree in the Department of Electrical Engineering and Computer Science at Syracuse University. Her research interests include video understanding, image segmentation, and object detection in computer vision, specifically activity recognition from wearable cameras.
\end{IEEEbiography}

\begin{IEEEbiography}[{\includegraphics[width=1in,height=1.25in,clip,keepaspectratio]{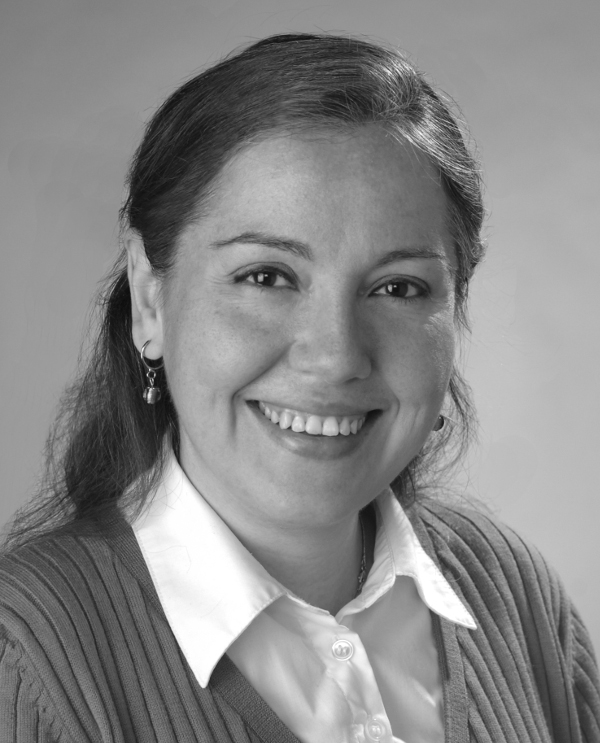}}]{Senem Velipasalar}

(M'04--SM'14) received the Ph.D. and M.A degrees in electrical engineering from Princeton University, Princeton, NJ, USA, in 2007 and 2004, respectively, the M.S. degree in electrical sciences and computer engineering from Brown University, Providence, RI, USA, in 2001, and the B.S. degree in electrical and electronic engineering from Bogazici University, Istanbul, Turkey, in 1999. From 2007 to 2011, she was an Assistant Professor with the Department of Electrical Engineering, University of Nebraska-Lincoln. She is currently a Professor in the Department of Electrical Engineering and Computer Science, Syracuse University. 
The focus of her research has been on machine learning, mobile camera applications, wireless embedded smart cameras, multicamera tracking and surveillance systems. 
She is a member of the Editorial Board of the IEEE Transactions on Image Processing and Springer Journal of Signal Processing Systems.
\end{IEEEbiography}

\vspace{11pt}


\vfill

\end{document}

%% file: 0_abstract.tex
Capturing interaction of hands with objects is important to autonomously detect human actions from egocentric videos. In this work, we present a pyramid video transformer with a dynamic class token generator for egocentric action recognition. Different from previous video transformers, which use the same static embedding as the class token for diverse inputs, we propose a dynamic class token generator that produces a class token for each input video by analyzing the hand-object interaction and the related motion information. The dynamic class token can diffuse such information to the entire model by communicating with other informative tokens in the subsequent transformer layers. With the dynamic class token, dissimilarity between videos can be more prominent, which helps the model distinguish various inputs. In addition, traditional video transformers explore temporal features globally, which requires large amounts of computation. However, egocentric videos often have a large amount of background scene transition, which causes discontinuities across distant frames. In this case, blindly reducing the temporal sampling rate will risk losing crucial information. Hence, we also propose a pyramid architecture to hierarchically process the video from short-term high rate to long-term low rate. With the proposed architecture, we significantly reduce the computational cost as well as the memory requirement without sacrificing from the model performance. We perform comparisons with different baseline video transformers on the EPIC-KITCHENS-100 and EGTEA Gaze+ datasets. Both quantitative and qualitative results show that the proposed model can efficiently improve the performance for egocentric action recognition.

%% file: 1_introduction.tex
\begin{figure}[h!]
  \centering
    \includegraphics[width=1\linewidth]{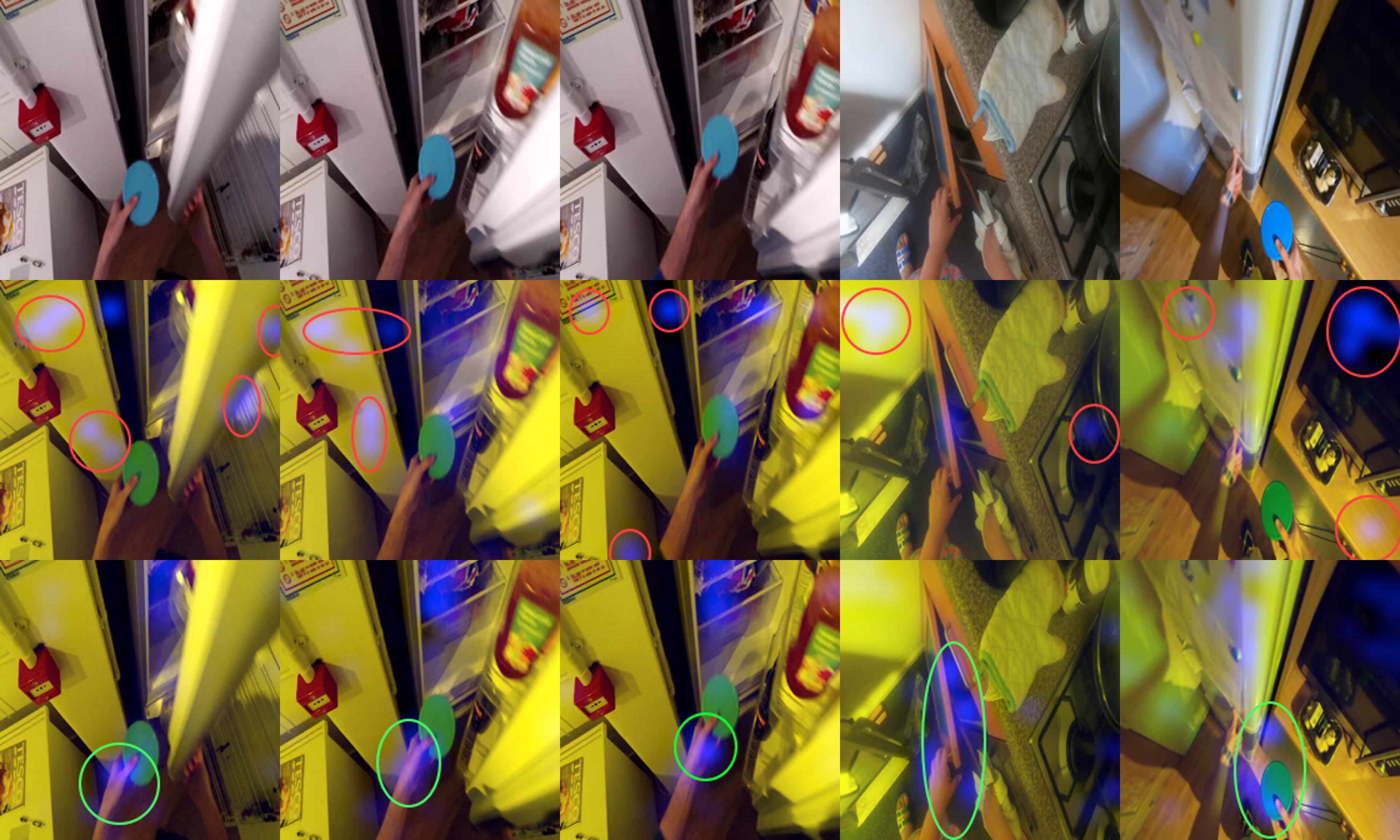}
  \caption{\textbf{Visualization of attention distribution.} Top to bottom row: RGB frames from original video, the frames with the attention map generated by TimeSformer and TimeSformer equipped with the proposed Dynamic Class Token Generator (DCTG), respectively.}
  \label{fig:DCTG_attention}
\end{figure}

\IEEEPARstart{L}{arge-scale} scene changes and fast camera motion make egocentric action recognition (EAR) a challenging problem in computer vision.~Egocentric view captures objects as well as the interactions between objects and the subject (see Fig.~\ref{fig:DCTG_attention}). Although transformer-based models have been introduced as powerful tools in video understanding area and achieved state-of-the-art (SOTA) performance~\cite{arnab2021vivit, bertasius2021space, liu2021video, patrick2021keeping, herzig2021object, li2021trear}, only a handful of them specifically considers the properties of the first-person view. In our work, we focus on improving the performance of video transformers for EAR. More specifically, we propose a novel framework, referred to as the EgoViT, which carefully takes the special properties of egocentric data into account, and can be integrated with different video transformers.  

For distinct input video clips, previous transformers use the same class token (CT), with other informative tokens, for  self-attention calculation. With fixed query (Q), key (K), and value (V) matrices during the inference time, at the first attention layer, the CT will always produce the same Q, K, and V vectors for any input. Although this kind of static CT can be assigned to various semantic messages, based on the input clip after passing through several model layers, it will cause the same component to be injected into the tokens of different video clips through the self-attention mechanism, which will likely weaken the ability to distinguish different videos. Thus, considering that CT can spread messages into other embeddings throughout the model, we argue that it can be more effective to equip the CT with useful action-related information.

The hand is the key to how humans interact with the world \cite{shan2020understanding}.~The interaction between hand(s) and object(s) is a main component of egocentric actions, as seen in Fig.~\ref{fig:DCTG_attention}, and precisely capturing the motion and appearance information of such interaction is a key factor for EAR. The relatively large background scene in egocentric videos is less important to determine actions, and also makes it harder for the algorithm to focus on the action-informative parts. Hence, directing the model to concentrate more on hand-object parts is especially important for EAR, which is not considered by current transformers.~Considering this, and to make the CT more functional, we propose a dynamic CT mechanism to generate a specific CT, enhanced with hand-object interaction information, for each input based on the content of the video clip, so as to guide the transformer to better recognize the egocentric action.~ORViT~\cite{herzig2021object} uses an object detector pretrained on MS COCO dataset as the off-the-shelf model to provide object information. Yet, the object classes in MS COCO are very different from the objects in egocentric datasets, which may confuse the model during EAR. Moreover, not all objects in the frames contribute to the human action, e.g. some objects are only part of the background.~Also, hand information is not considered by ORViT. We have performed experiments to compare the objects information and hand-object information (that we use), and the results show that hand-object information is more useful for EAR. This experiment is included in the supplementary material. We also compare our approach with ORViT in terms of performance and model size in Sec.~\ref{sec:4_exp}.

The time axis is a special dimension for videos. Most of the video transformers extract temporal semantic information globally from the entire video, ignoring the large scene variation (due to large camera movement) in egocentric videos. Although Swin~\cite{liu2021video} uses two non-overlapping windows along the time axis, there is no mechanism to capture the hierarchical temporal feature of a video, and there is still a lot of repeated scene information within a window. Due to this, most pair-wise calculations between frames generate redundant information. To make it more clear, in Fig.~\ref{fig:t_vec}, we show the feature vectors of each temporal space from the last layers of TimeSformer~\cite{bertasius2021space} and Swin. Both models have the 32-frame input, and at the last layer, the number of features along the time axis are 32 and 16 for TimeSformer and Swin, respectively. We apply Principal Component Analysis (PCA) to project the high-dimensional temporal features to 3D space, and the number marker corresponds to the temporal position of each vector.~It can be seen that, for both models, features from close time instances are also close to each other in the feature space, and sometimes even overlap (marked with red circle). The features form several clusters in the space, indicating the high similarity of the information provided by nearby frames. This further proves that, after the whole model processing, there still exists a lot of redundant information in the neighboring frame vectors. Therefore, we argue that, with large scene changes, egocentric videos can always be decomposed into several phases. For example, the action ``take pasta container" shown in Fig.~\ref{fig:phase_attn}, having a duration of about 1.8s, can be roughly divided into four phases. In the first phase, the person is looking for and approaching the container. The background shows a kitchen counter with the container in view. In the second phase, the person is holding the container, and the background is almost fixed. In the third phase, after getting the container, the person is turning around.~The background changes greatly, while only half of the container is in the view. In the fourth phase, the container is not in the view, and the background becomes another side of the kitchen, which is totally different from the initial one.~Thus, to describe the video more comprehensively and decrease the computation cost, we propose a pyramid architecture.~In this example, the scenes of the first and second phases basically overlap, and the target object remains in the field of view. Starting from the third phase, the scene undergoes a large change, and the target object starts to move out of view and disappears completely in the last phase. Thus, the frames of the fourth phase are less important for action recognition. However, existing transformers still arrange all the frames to do calculations with each frame in the fourth phase. The average pooling approach in the last stage of Swin Video transformer averages the contribution of all frames, and does not emphasize important frames.~Hence, considering the uneven contributions of the phases, we also propose a dynamic merging mechanism to adapt to our pyramid structure and enhance the class token. 

\textbf{Contributions.} The main contributions of this work include the following: (i) We propose a novel video transformer, EgoViT, for EAR. EgoViT takes into account the hand-object information, which is not considered by previous transformers; (ii) We propose a dynamic class token generator (DCTG) producing the CT, carrying EAR-related information based on the content of each input, instead of the static CT used by previous transformers. Dynamic CT can guide the model to focus on informative parts of the video; (iii) Considering the large camera movement in egocentric videos, we construct the transformer by a Pyramid Architecture with a Dynamic Merging (PADM) module, which can properly model the temporal structure and dynamically distribute weights to each temporal component; (iv) We conduct extensive experiments to show that our proposed EgoViT can boost the performance of various transformers for EAR while reducing the computations at the same time.

%% file: 2_related_work.tex
\textbf{Transformers in Video Recognition:} The self-attention model proposed by Vaswani et al.~\cite{vaswani2017attention} replaces the CNN or RNN layers with self-attention layers, and was a big success in the natural language processing area. More recently, Dosovitskiy et al.~\cite{dosovitskiy2020image} proposed a pure Vision Transformer (VIT) for the image classification task by taking advantage of a super large 300M JFT dataset~\cite{sun2017revisiting}.~Many other works have focused on building vision transformer models with lower computational cost by using different strategies, such as using semantic visual tokens~\cite{xie2021so}, layer-wise token to token transformation~\cite{yuan2021tokens}, adding distillation losses~\cite{touvron2021training} and building a hierarchical structure with the shifted windows~\cite{liu2021swin}.~Video transformers~\cite{Neimark2021VTN, arnab2021vivit,bertasius2021space, liu2021video, patrick2021keeping, herzig2021object} have mirrored the advances in image understanding and achieved SOTA performance on the major video recognition benchmarks~\cite{kay2017kinetics,goyal2017something}.~Many of previous video transformers~\cite{bertasius2021space, arnab2021vivit, Neimark2021VTN} simply extend the image spatial domain to the global temporal/spatiotemporal domain, which leads to high computation costs, and the performance heavily depends on the 2D spatial model pre-trained on super large datasets JFT-300M~\cite{sun2017revisiting} or ImageNet-21k~\cite{Russakovsky15imagenet}. To reduce the computation and memory costs as well as provide locality inductive bias in the self-attention module, Liu et al.~\cite{liu2021video} strictly followed the hierarchy of the original Swin Transformer~\cite{liu2021swin} for the image domain, and extended the scope of local attention computation from only the spatial domain to the spatiotemporal domain. However, the global temporal self-attention cannot be considered by simply using the shifted windows mechanism. We argue that both local and global temporal attention are critical, especially for egocentric videos, which are usually captured by wearable cameras with large and frequent movements. Our proposed hierarchical pyramid structure successfully provides an inductive bias on grouping the local temporal attentions as well as the high-level global temporal attentions, which can successfully handle the camera motion across different scenes.

\textbf{Object detection-orientated video action recognition:} Object-human/object-hand interaction models~\cite{shan2020understanding, fouhey2018lifestyle, rogez2015understanding, gkioxari2018detecting, wang2021discovering} have been widely explored and achieved significant success. Given that object-human interaction is a key feature for the video action recognition task, many existing models~\cite{kato2018compositional, gao2020drg, xu2019learning, wang2020symbiotic, baradel2018object} employed object detection and interaction features for video understanding. Herzig et al.~\cite{herzig2021object} designed an ``Object-Dynamics Module", which can be inserted into any transformer model, and achieved SOTA performance in video action recognition. This work is the most related one to ours, however, it is pointed out in \cite{herzig2021object} that the improvement on the egocentric videos, such as EPIC-KITCHEN100 dataset, is not as impressive as other datasets because of the frequent and large camera movement. We believe that the major reasons are 1) object-subject interaction features are not being considered in~\cite{herzig2021object}; 2) locality inductive bias is not provided in the self-attention module. To the best of our knowledge, our proposed method is the first attempt to inject the object-human interaction features into the transformer models by designing a dynamic class token, and dynamically embedding the object-human interaction features into the class token.

\textbf{Egocentric video action recognition:} Thanks to the increasing availability of wearable cameras and several egocentric video datasets~\cite{Damen2021PAMI,damen2018scaling,sigurdsson2018charades,li2018eye}, the research in egocentric video analysis has made significant strides.~The general video transformer models may not work for egocentric videos because of the frequent and large camera movements as well as the complicated background scene. Herzi et al.~\cite{herzig2021object} proposed an object centric module that can be plugged into video transformer models. Wang et al.~\cite{wang2020symbiotic} designed a symbiotic attention with object-centric feature alignment framework to provide reasoning between the actor and the objects. Huang et al.~\cite{huang2021towards} provided some effective training strategies for the general transformer models on Epic-Kitchens dataset. Although the aforementioned attempts provided promising results on Epic-Kitchens dataset, they did not, in general, focus on addressing the specific challenges existing in egocentric videos. In this paper, we propose a pyramid video transformer structure, with dynamic class token, which is shown to successfully address both of these challenges with egocentric videos. 

%% file: 3_model.tex
\begin{figure*}
  \centering
    \includegraphics[width=0.75\linewidth]{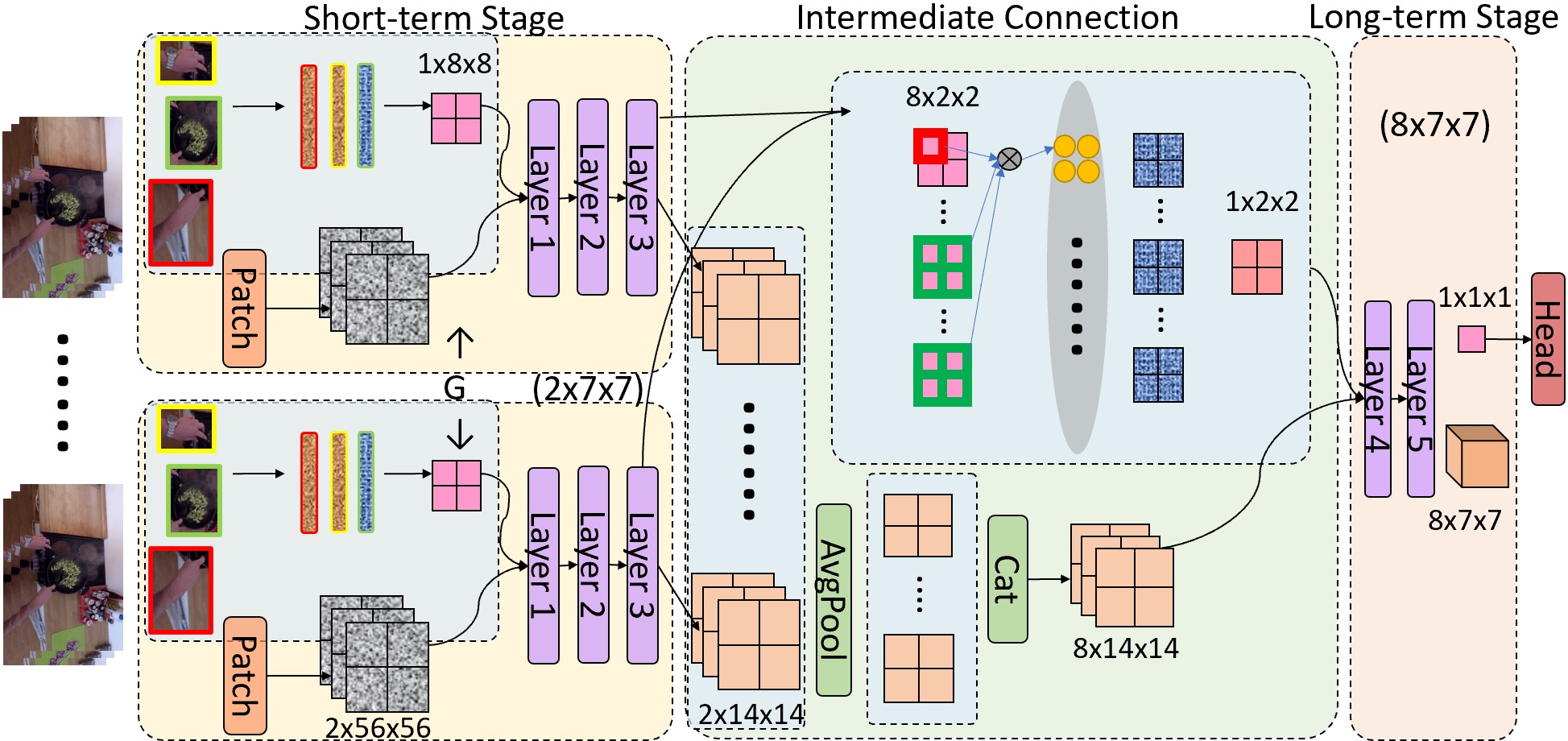}
    \caption{
    The structure of the proposed EgoViT with a pyramid architecture incorporating Dynamic Class Generator.}
    \label{fig:model}
\end{figure*}

\begin{figure}
   \centering
\includegraphics[width=0.9\linewidth]{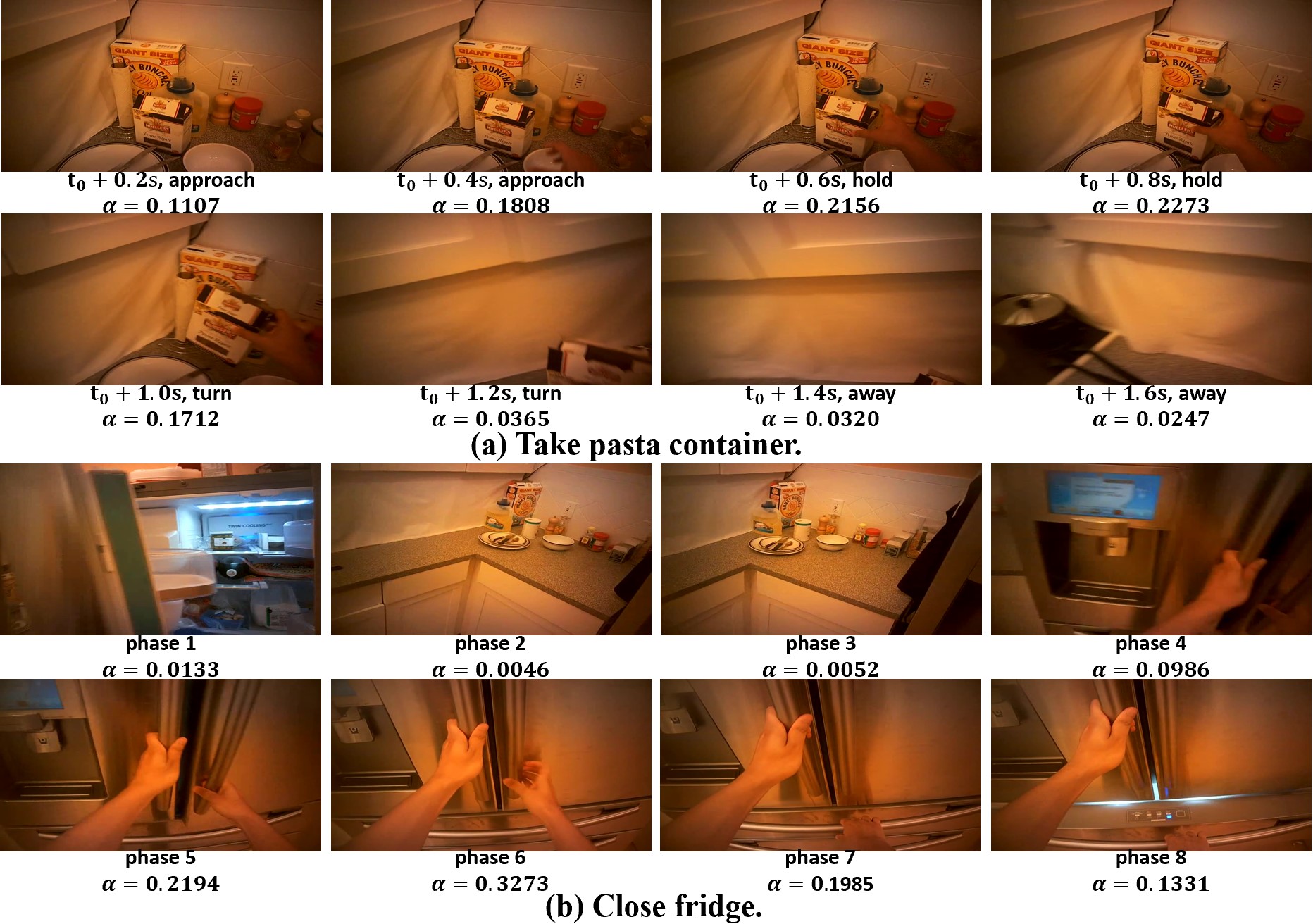}
     \caption{Visualization of phase attention distribution.}
   \label{fig:phase_attn}
\end{figure}
We first provide a summary of transformers in Sec.~\ref{ssec:prelim}. We introduce the dynamic class token module for global and local transformer in Sec.~\ref{ssec:token}, and then present the pyramid architecture
in Sec.~\ref{ssec:pyramid}.

\subsection{Preliminaries}
\label{ssec:prelim}
The pure self-attention based transformer~\cite{vaswani2017attention} is mainly used as a sequence transduction model. Vision transformers~\cite{dosovitskiy2020image,arnab2021vivit} convert images/video frames into serialized data to be able to process them by a transformer. Let $I \in \mathbb{R}^{T \times H \times W \times C}$ denote the input video clip with $T$ image frames of size $H \times W \times C$.~$I$ is first segmented into non-overlapping image patches of size $\mathbb{R}^{ P_{T} \times P_{H} \times P_{W} \times C}$. Then, these patches are flattened into sequence data $\mathbb{R}^{N_{P} \times (P_{T} \times P_{H} \times P_{W} \times C)}$, where $N_{P} = T_{P} \times H_{P} \times W_{P} = \frac{T}{P_{T}} \times \frac{H}{P_{H}} \times \frac{W}{P_{W}}$. A matrix of size $\mathbb{R}^{(P_{T} \times P_{H} \times P_{W} \times C) \times D }$ is used to linearly map the patch sequence into a $D$-dimensional space.~The generated 
vectors $\hat{X}_{P} \in  \mathbb{R}^{N_{P} \times D}$ are treated as the video embeddings and the main inputs of the subsequent transformer blocks.

3D video data is highly structured having strict spatial relation and temporal order. Since converting 3D video to 1D sequence
causes losing these relations, it is necessary to preserve the 3D position information in the video embeddings at the very beginning. Vision transformer usually initializes a position embedding $(x_{pos})_{a,b} \in \mathbb{R}^{D}$ and a temporal embedding $(x_{temp})_{t} \in \mathbb{R}^{D}$, for the corresponding video embedding $(\hat{x}_{P})_{t,a,b}$ at the 3D position $(t,a,b)$, to learn the position and temporal bias during training, as shown in Eq.~(\ref{eq:x_p}):
\begin{equation}\label{eq:x_p}
    (x_{P})_{t,a,b} = (\hat{x}_{P})_{t,a,b} + (x_{temp})_{t} + (x_{pos})_{a,b} ,
\end{equation}
where $t \in [1,T_{P}]$, $a \in [1,H_{P}]$, and $b \in [1,W_{P}]$.

In the global-attention based transformers \cite{bertasius2021space,patrick2021keeping,herzig2021object}, in addition to the video embeddings, there is usually a classification token (CT), $x_{cls} \in \mathbb{R}^{D}$, which is initialized to be concatenated with other informative tokens, and participates the self-attention calculation like other tokens in the subsequent transformer blocks. This special token is used as the token for the final classification. Thus, the input of the transformer block is the concatenation of the class token and the informative video embeddings, as shown in Eq.~(\ref{eq:x}),
\begin{equation}\label{eq:x}
\small{
    x = [x_{cls}, (x_{P})_{:}],
    }
\end{equation}
where subscript $:$ is used to denote a full slice of the input tensor in that dimension.

Swin video transformer~\cite{liu2021video}, which is a local-window self-attention based transformer, applies 3D average pooling among the output embeddings from the last transformer block, instead of using a classification token, to produce the feature vector for final classification.

\subsection{Dynamic Class Token Generator} \label{ssec:token}
For EAR, the appearance and dynamic interaction of hands and objects are the key clues to understand human actions~\cite{shan2020understanding}. Hence, we argue that exploring such information and circulating it from the beginning to the end of the model can encourage the transformer to focus more on the action-informative parts of egocentric videos. Considering this, we propose a dynamic class token generator (DCTG), with negligible number of parameters, to guide the transformer by a class token enhanced with hand-object interaction information.

We first apply a pre-trained hand-object detector (HOD)~\cite{shan2020understanding}, which is built based on an object detection system, more specifically Faster-RCNN~\cite{ren2016faster}. We choose HOD as the offline hand-object feature extractor for the following reasons: (i) HOD is specifically trained to identify two objects -- human hands and contacted objects; (ii) the model has been shown to generalize well across datasets especially egocentric datasets; (iii) official datasets like EPIC-KITCHENS-100 \cite{damen2020rescaling} provide automatic hand-object detections based on HOD, indicating the model's reputation and reliability. We send $T$-many frames $I_{t}$ to HOD to get bounding box predictions $BB_{t}$ for hands and objects, as well as the feature maps $I_{t}^{base}$ generated by the base part of HOD. According to the credibility ranking, we choose the top-M hand and top-M object detections with confidence score $\theta > 0.5$, and resend them, along with the feature maps, to `ROIAlign' module and the `top feature refine module' of HOD to obtain the final 2048-D feature vector for each selected detection. Finally, we concatenate them together to get the hand-object (HO) feature $F^{HO} \in \mathbb{R}^{T \times (2M) \times 2048}$ for each video as the input of the DCTG. The procedure is shown in Eq.~(\ref{eq:hod}):
\begin{equation}\label{eq:hod}
\begin{aligned}
&cls_{t},BB_{t},\theta_{t}  = \text{HOD}(I_{t}), \,\,\text{for   } t \in [1,T]\\
&I_{t}^{base} = \text{HOD}_{base}( I_{t}) \in \mathbb{R}^{1024 \times H^{b} \times W^{b}}, \\
    &I_{t}^{align} = \text{ROIAlign}(I_{t}^{base},  BB_{t}) \in \mathbb{R}^{2M \times 1024 \times H^{a} \times W^{a}}, \\
&    I_{t}^{HO} = \text{HOD}_{top}( I_{t}^{align}) \in \mathbb{R}^{2M \times 2048}, \\
 &   F^{HO} = [I_{1}^{HO}, ..., I_{T}^{HO} ]
\end{aligned}
\end{equation}

Then, our goal is to explore the dynamic hand-object interaction, and utilize such knowledge to generate the informative CT for each egocentric video clips. To study the inter-feature relationship and integrate the $2M$ features into one feature vector representing the hand-object information for each frame, we attempt (a) applying query-key-value (QKV) self-attention feature-wise, then averaging along the feature axis; (b) directly averaging the $2M$ features, to get the T-frame hand-object feature $F^{HO'} \in \mathbb{R}^{T \times 2048}$. To explore the inter-frame communication and unearth the hand-object dynamic clues, we experiment with two typical methods for aggregating knowledge in temporal space: (1) QKV-self-attention; and (2) long short term memory (lstm). Then, the specific class token is produced by either calculating the average frame vector from (1) or picking out the last output state vector from (2). While exploring the temporal information from features, we keep the tensor flow in the same dimension space as the video embeddings to produce the class token $x_{cls} \in \mathbb{R}^{D}$. We conduct ablation studies (results are provided in the supplementary materials) on the four combinations of the above mentioned methods, and conclude that directly averaging the $2M$ features for each frame, and then applying two lstm layers---the combination of (b) and (2)---is the best scheme. Therefore, the procedure of dealing with hand-object features in DCTG can be expressed as in Eq.~(\ref{eq:DCTG}).

\begin{equation}\label{eq:DCTG}
\small{
\begin{aligned}
    F_{t}^{HO'} = \frac{1}{2M}\sum_{i}^{2M}F_{t,i}^{HO}; \quad & t \in [1,T], i \in [1,2M], \\
    F^{HO''} = \text{LSTM}([F_{1}^{HO'}& ,...,F_{T}^{HO'} ]) \in \mathbb{R}^{T \times D}, \\
    x_{cls} =F_{T}^{HO''} & \in \mathbb{R}^{D}.
\end{aligned}
}
\end{equation}

For the global self-attention based transformer \cite{bertasius2021space,arnab2021vivit,patrick2021keeping}, there is a single CT initialized for the whole input clip at the beginning, so we directly replace it with our dynamic CT. For the local-window self-attention based transformer \cite{liu2021video}---swin video transformer, we generate a CT map. The swin video transformer contains 3D window-based multi-head self-attention module. Assume the size of the 3D local window $WI$ is $\mathbb{R}^{WI_{T} \times WI_{H} \times WI_{W}}$, then for the patched input $X^{P} \in \mathbb{R}^{ (T_{P} \times H_{P} \times W_{P}) \times D} $, there will be $N_{WI} = T_{WI} \times H_{WI} \times W_{WI}$ windows, where $T_{WI} = \frac{T_{P}}{WI_{T}}$, $H_{WI} = \frac{H_{P}}{WI_{H}}$, and $W_{WI} = \frac{W_{P}}{WI_{W}}$. We initially assign the same dynamic CT to all windows, thus forming a class token map $(X^{cls})^{0} \in  \mathbb{R}^{(T_{WI} \times H_{WI} \times W_{WI}) \times D}$. Then through the subsequent layers, each window updates its own CT. The operation in the window-based multi-head self-attention with dynamic CT is expressed in Eq.~(\ref{eq:swin_cls}):
\begin{equation}\label{eq:swin_cls}
\small{
\begin{aligned}
    (I_{WI})_{i} & = [(x_{cls})_{i}, (x_{WI})_{i,:}]; i \in [1,N_{WI}], \\
    (\hat{I}_{WI})_{i}^{l} & = \mathit{\text{W-MSA}(\text{LN}((I_{WI})_{i}^{l-1})) + (I_{WI})_{i}^{l-1} }, \\
    (I_{WI})_{i}^{l} & = \mathit{\text{MLP}(\text{LN}((\hat{I}_{WI})_{i}^{l})) + (\hat{I}_{WI})_{i}^{l} }, \\
    (\hat{I}_{WI})_{i}^{l+1} & = \mathit{\text{SW-MSA}(\text{LN}((I_{WI})_{i}^{l})) + (I_{WI})_{i}^{l} }, \\
    (I_{WI})_{i}^{l+1} & = \mathit{\text{MLP}(\text{LN}((\hat{I}_{WI})_{i}^{l+1})) + (\hat{I}_{WI})_{i}^{l+1} }, \\
\end{aligned}
}
\end{equation}
where $LN$ refers to the LayerNorm operation, $(I_{WI})_{i} \in \mathbb{R}^{(1+WI_{T} \times WI_{H} \times WI_{W})\times D}$ denotes the tensor of the $i^{th}$ window, (S)W-MSA represents the (shifted) window-based self-attention module, $(\hat{I}_{WI})_{i}^{l}$ and $(I_{WI})_{i}^{l}$ represent the $i^{th}$ window output tensor of (S)W-MSA and MLP at the $l^{th}$ transformer block, respectively. Then, to adapt to the hierarchical framework, we merge the class token map at the same time as the patch merging of informative tokens. We concatenate the class tokens within the $2\times2$ spatial neighborhood, which makes the shape of the class token map $\mathbb{R}^{(T_{WI} \times \frac{H_{WI}}{2}  \times  \frac{W_{WI}}{2} ) \times 4D}$. Then, we apply a linear matrix with shape $\mathbb{R}^{ 4D \times 2D}$ to map the $4D$ concatenated tokens to $2D$ space. Therefore, this step reduces the spatial size and increases the its dimension by 2 times for the class token map. We apply this step at the end of stage 1, stage 2, and stage 3, which keeps the same patch merging scheme of the informative tokens in swin. The output class token map of the last stage (stage 4) will be in $\mathbb{R}^{(T_{WI} \times \frac{H_{WI}}{8}  \times  \frac{W_{WI}}{8} ) \times 8D}$. Instead of using the 3D average pooling on the patch tokens map as in the original swin, we apply 3D average pooling among the dynamic class token map to obtain a single $8D$ vector as the final feature for action prediction. 

Therefore, through our DCTG, the model will be more inclined to focus on and extract features related to egocentric actions in the subsequent attention mechanism. In addition, our DCTG adds only negligible number of parameters to the model. In order to motivate and validate the merit of the dynamic class token, we conduct an evaluation in Sec.~\ref{ssec:Qualitative} comparing the visualizations of dynamic class token and static class token based on the attention weights for different image components. The results show that our dynamic class token is more conducive for the model to focus on the core part of the egocentric actions.

\subsection{Pyramid Architecture with Dynamic Merging} \label{ssec:pyramid}
Our pyramid architecture is composed of two sequential stages, which operate at two different frame rates, and are connected by our proposed dynamic merging mechanism. Since other global attention-based transformers have a similar structure as Swin, we use Swin as our base model to illustrate our framework, as shown in Fig.~\ref{fig:model}, without loss of generality.

In the first stage, the model focuses on exploring the intra-frame relationship for short-term actions of the subject by separating the video into $G$ phases. In other words, for $T$ input frames, we divide them into $G$ groups with $\frac{T}{G}$ consecutive frames in each group. Thus, each phase has the same frame rate as the raw input clip. We send $\frac{T}{G}$ frames to the DCTG, Patch Embedding, and $L1$ layers of transformer blocks. The temporal window size of each group is set to $\frac{T_{p}}{G}$, and the weights of these modules are shared among the groups. The goal of this stage is to perceive the actions under the almost fixed scene. After the processing, there will be class tokens $x_{cls}^{S1} \in \mathbb{R}^{G \times H_{wi}^{s1} \times W_{wi}^{s1} \times D^{s1} } $ and normal tokens $x_{p}^{s1} \in \mathbb{R}^{G \times \frac{T}{G} \times H^{s1} \times W^{s1} \times D^{s1}} $ from the $G$ phases. We perform average pooling along the temporal axis in each group and then concatenate the pooled tensors to obtain the input normal tokens for the second stage.~The class token can be regarded as a summary of the semantic meaning of each short-term action.~Since the contribution of each phase is not fixed for various videos, we propose a dynamic merging module to assign weights and aggregate the CTs from the first stage. The goal of dynamic merging is to assign a larger weight to the CTs representing the key short-term actions, like ``approach pasta-container" and ``hold pasta-container" in the ``take pasta-container" example. These key actions have similar scene in the background, and we assume that their class tokens have similar directions.
We first obtain the score $\alpha _{g,s,g'}$ of CT $\mathbf{x}^{cls}_{g,s}$ at different groups $g'$ by calculating the dot product between it and the CT at $g'$ group then dividing it by the product of their l2 norms. Then, we sum the scores along spatial and group axis to get the total score $\alpha _{g,s}$ for $(\mathbf{x}_{cls})_{g,s}$. We normalize the scores for all class tokens along group axis by softmax operator, and the final class token map for the second stage is obtained by the weighted sum of class tokens along group axis.
The procedure can be expressed in Eq.~(\ref{eq:dynamic_merging}):
\begin{equation}\label{eq:dynamic_merging}
\small{
\begin{aligned}
    \alpha _{g,s,g'} &= \frac{1}{\left \| \mathbf{x}^{cls}_{g,s} \right \|}\sum_{s'}\frac{\mathbf{x}^{cls}_{g,s} \cdot \mathbf{x}^{cls}_{g',s'}}{\left \| \mathbf{x}^{cls}_{g',s'} \right \|} ,\\
    \alpha _{g,s} &= \sum _{g'\neq g}\alpha _{g,s,g'} ,\\
    \mathbf{x}^{cls}_{s} &= \sum _{g}\mathbf{x}^{cls}_{g,s}\cdot \frac{exp(\alpha _{g,s})}{\Sigma _{\bar{g}}exp(\alpha _{\bar{g},s})} ,\\
\end{aligned}
}
\end{equation}
where $\mathbf{x}^{cls}_{s}$ is the final merged class token map from the first stage. In this way, we dynamically tune the weights for each group.

Therefore, after the merging module between the short-term and long-term stages, the model has an informative token map $x_{info}^{s1} \in \mathbb{R}^{G \times H^{s1} \times W^{s1} \times D^{s1} } $ gathering the information of short-term actions, and the intermediate class token map $x_{cls}^{s1} \in  \mathbb{R}^{1 \times H_{wi}^{s1} \times W_{wi}^{s1} \times D^{s1} } $ inferred from the short-term actions, where $s1$ indicates the first stage. The goal of the second stage, or long-term stage, is to perceive the action in long duration under large-scale scene changes by exploring the inter-relationships of the short-term actions. Therefore, with the combined token maps, we design the subsequent blocks to have a global view on the time axis. At the end of this stage, we have informative token map $x_{info}^{s2} \in \mathbb{R}^{G \times H^{s2} \times W^{s2} \times D^{s2} } $ and the final class token map $x_{cls}^{s2} \in  \mathbb{R}^{1 \times H_{wi}^{s2} \times W_{wi}^{s2} \times D^{s2} } $, where $s2$ indicates the second stage. Then, we apply average pooling on class token map among the spatial axes to get a single $8D^{s2}$ feature vector, and we send it to head for the final classification.

Perceiving the video from local to global view, the processed class token meticulously collects information on short-term actions, gives prominence to critical phases, and explores long-term features. In this way, we have a comprehensive understanding of the video. In summary, our model takes into account three characteristics of egocentric video, namely large-scale scene changes between far frames, high overlap between near frames, and different contributions of phases, so that it can avoid redundant information, reduce the impact of less important frames, focus on more important potions, and also decrease the amount of parameters and computation.

%% file: 4_experiments.tex
\textbf{Datasets}. 
\textbf{EPIC-KITCHENS-100} \cite{damen2020epic} (EK100) is the largest dataset in first-person (egocentric) vision, capturing different activities in a kitchen over multiple days. There are 90K action segments, 97 verb classes, and 300 noun classes in the dataset. These action instances follow a long-tailed distribution. Some largest classes (i.e. those with most instances), which we call the many-shot classes, account for 80\% of the total number of instances in the dataset. We additionally split two subsets, containing only the many-shot verb classes and the many-shot noun classes, for our ablation study. In our experiments, the models for verb prediction and noun prediction are trained separately. Most of the clip lengths of this dataset are distributed around 128 frames, so we keep the sample duration as 128 frames, that is, 32-frames sampled with 4-frame interval and 64-frame sampled with 2-frame interval.

\textbf{EGTEA Gaze+} (EGaze+)\cite{li2018eye} is the Extended GTEA Gaze+ dataset. It contains 29 hours of first person videos from 86 unique sessions. In these sessions, 32 subjects perform 7 different meal preparation tasks in a naturalistic kitchen environment. The dataset also comes with action annotations of 10321 instances from 106 classes with an average duration of 3.2 sec, at a frame rate of 24FPS. In our experiments, we sample 32 frames from each clip with a 2-frame stride. There are three training/testing splits provided with the official dataset, and we use the first split (8299 for training, 2022 for testing) to evaluate the performance of action recognition.

\subsection{Ablation Study on Group and Stage Depth}
\label{ssec:ablation}
To prove the effectiveness of the proposed DCTG and study the hyper parameters for the PADM module, namely the number of phases G and the depth ratio (DR) of the two stages, we conduct experiments on EK100 with TimeSformer as our baseline model. All the models are initialized by the TimeSformer pre-trained on the Something-Something V2~\cite{goyal2017something} dataset.

We first compare the performance of TimeSformer with and without the proposed DCTG module. As seen in Tab.~\ref{tab:ablation}, the model is consistently improved across all columns. More specifically, the model is improved by 2.45\%/4.5\%, 2.48\%/1.84\%, 3.72\%/0.49\%, and 3.16\%/1.14\% on many-shot verb, many-shot noun, verb, and noun, respectively, with 32/64 frames sampling, which indicates that the proposed DCTG is effective in improving model performance.

As for the number of phases $G$, we compared 4-phase and 8-phase settings. From the Tab.~\ref{tab:ablation}, we can see that, with the same $DR$ value, 8-phase model always performs better than 4-phase model, which further proves the importance of action decomposition for EAR. For the depth ratio (the depth of stage 1 / the depth of stage 2), we compared the values of 0.5, 1, and 2. As shown in Tab.~\ref{tab:ablation}, in most cases, $DR=2$ produces best results, except for the $G=4$ case in many-shot verb subset. We reason that the 4-phase model cannot partition the verb (motion) of the action very well, so that the deeper the stage 1 is, the worse the model learned.

In summary, with both the DCTG module and proper PADM settings, the TimeSformer is significantly improved for all cases. Therefore, we employ the $G=8$ and $DR=2$ as the setting of PADM for the remaining experiments. Since there is not much difference in performance when using 64 or 32 frames, we only sampled 32 frames in the following experiments.

\begin{table}
\begin{center}
\caption{\textbf{Ablation study with TimeSformer on EPIC-KITCHENS-100 (EK100) and its many-shot subsets}. V and N indicate the prediction accuracy for verb class and noun class, resp., and MV and MN indicate the prediction accuracy on the many-shot verb subset and many-shot noun subset, respectively. We use ``Ours" to represent the model equipped with both DCTG and PADM modules.}
\label{tab:ablation}
\begin{tabular}{|l|c|c|c|c|c|c|}
        \hline
        Models          & T                    & G/R & MV & V  & MN & N  \\
        \hline
        TSformer & 32                      &        -       & 68        & 57.85 & 47.09     & 41.75 \\
        \hline
        +DCTG & 32                 &         -      & \textbf{70.45}     & \textbf{61.57} & \textbf{49.57}    & \textbf{44.91} \\
        \hline
        \multirow{4}{*}{+PADM} & 32  & 4/1       & 65.69     &   -    & 47.89     &   -    \\
                                        & 32    & 8/1       & 68.66     &    -   & 48.67     &    -   \\
                                        & 32    & 8/0.5       & 68.26     &   -    & 48.02     &   -    \\
                                        & 32    & 8/2       & \textbf{71.79}     & \textbf{62.33} & \textbf{50.8}     & \textbf{44.86} \\
        \hline
        \multirow{4}{*}{Ours} & 32 & 4/1       & 69.54     &    -   & 49.32     &    -   \\
                                        & 32    & 4/2       & 69.49     &    -   & 49.88     &    -   \\ 
                                        & 32    & 8/1       & 71.45     &    -   & 50.04     &    -   \\ 
                                        & 32    & 8/2       & \textbf{72.26}     & \textbf{62.42} & \textbf{50.88}     & \textbf{45.76} \\
        \hline
        TSformer                  & 64    &         -      & 65.2      & 60.21 & 48.02     & 42.36 \\
        +DCTG             & 64    &       -        & 69.7      & 60.7  & 49.86     & 43.5  \\
        +PADM               & 64    & 8/2       & 68.04     & 62.35 & 49.7      & 44.97 \\
        Ours             & 64    & 8/2      & \textbf{72.39}     & \textbf{62.38} & \textbf{51.25}     & \textbf{45.84} \\
        \hline
\end{tabular}
\end{center}
\end{table}

\subsection{Ablation Study on DCTG Module}
\label{sec:DCTG_Ablation_Study}
To investigate how to generate class tokens that can better direct the model, we conduct ablation studies on three aspects: (i) what type of off-the-shelf features can be more useful for EAR, (ii) how to explore inter-feature relationship to produce a single feature vector for each frame, and (iii) how to explore inter-frame relationship to generate a single class token for each input.
For the first aspect, we experiment with 2-class hand-object features extracted from the Hand-object detector \cite{shan2020understanding} pretrained on the 100DOH dataset \cite{shan2020understanding}, and 80-class object features extracted from the Mask-RCNN \cite{he2017mask} pretrained on MS COCO dataset \cite{lin2014microsoft}. 
For the second aspect, we try (a) applying query-key-value (QKV) self-attention feature-wise, then averaging along the feature axis; (b) directly averaging the 2M features.
For the third aspect, we experiment with two typical methods for aggregating knowledge in temporal space: QKV self-attention; and long short term memory (LSTM). 
Then, the specific class token is produced by either calculating the average frame vector from the former one or picking out the last output state vector from the latter one.

The results are shown in Table~\ref{tab:DCTG_Ablation_Study}. As can be seen,  directly averaging the hand-object features for each frame, and then applying LSTM layers provides the best performance.

\begin{table}
\begin{center}
\caption{\textbf{Ablation study with DCTG on EPIC-KITCHENS-100's many-shot subsets}. MV and MN indicate the prediction accuracy on the many-shot verb subset and many-shot noun subset, respectively.}
\label{tab:DCTG_Ablation_Study}
\begin{tabular}{|l|c|c|c|c|}
        \hline
        Feature type                 & Inter-feature        & Inter-frame & MV & MN \\
        \hline
        \multirow{4}{*}{Hand-object} & \multirow{2}{*}{QKV} & QKV & 56.24 & 36.62 \\
        \cline{3-5}
                                     &                      & LSTM & 56.64 & 36.88 \\
                                     \cline{2-5}
                                     & \multirow{2}{*}{Avg} & QKV & 57.12 & 38.31 \\
                                     \cline{3-5}
                                     &                      & LSTM & \textbf{60.81} & \textbf{38.45} \\
        \hline
        \multirow{4}{*}{Objects}     & \multirow{2}{*}{QKV} & QKV & 38.58 & 21.53 \\
        \cline{3-5}
                                     &                      & LSTM & 37.71 & 16.42 \\
                                     \cline{2-5}
                                     & \multirow{2}{*}{Avg} & QKV & 37.23 & 21.26 \\
                                     \cline{3-5}
                                     &                      & LSTM & 36.09 & 20.85 \\
        \hline
\end{tabular}
\end{center}
\end{table}

\subsection{Quantitative Results}
\label{ssec:Quantitative}
We compare our proposed EgoViT with the most well-known video transformers as well as several SOTA CNN-based models. The video transformers we compare with are TimeSformer (TSformer)~\cite{bertasius2021space}, Vivit~\cite{arnab2021vivit}, Motionformer (Mformer)~\cite{patrick2021keeping}, Swin~\cite{liu2021video}, and Object-Region Video Transformer (ORViT)~\cite{herzig2021object}. The number of training epochs, learning rate, and augmentation methods are kept the same as the corresponding transformer baselines. We set the number of hand-object features to 4 in EK100, and 2 in EGaze+. The results, summarized in Tables \ref{tab:epic100} and \ref{tab:egtea}, show that our proposed EgoViT outperforms all baseline transformers. More specifically, it improves TSformer by 2.65\%, 2.17\%,  3.48\%, and 1.1\%; improves Swin-S by 0.92\%, 1.54\%, 1.26\%, and 0.05\%;improves Swin-B by 0.2\%, 0.83\%, 0.1\%, and 0.47\%; improves Mformer-HR by 1.8\%, 2.85\%, 0.57\%, and 0.46\%, on Action (A), Verb (V), Noun (N) prediction from EK100, and top1 accuracy for EGAZE+, respectively.
\begin{table}
\begin{center}
\caption{\textbf{EAR results on the EPIC-KITCHENS-100}. First block presents the results of typical CNN-based models, while other blocks present the comparison results of previous transformers and our EgoViT-based transformers.}
\label{tab:epic100}
\begin{tabular}{|l|c|c|c|c|}
        \hline
        Models           & A     & V     & N     & pretrain \\ 
        \hline
        TSN \cite{damen2020rescaling}            & 33.57 & 60.2  & 46    & IN-1K    \\ 
        TRN \cite{damen2020rescaling}            & 35.28 & 65.9  & 45.4  & IN-1K    \\ 
        TBN \cite{damen2020rescaling}            & 35.55 & 66    & 47.2  & IN-1K    \\ 
        SlowFast \cite{feichtenhofer2019slowfast}     & 36.81 & 65.6  & 50    & K400     \\ 
        TSM \cite{damen2020rescaling}             & 37.39 & 67.9  & 49    & IN-1K    \\ 
        MBT \cite{nagrani2021attention}            & 43.4  & 64.8  & 58    &  -        \\ 
        TempAgg \cite{sener2021technical}       & 45.26 & 66    & 53.35 &  -       \\ 
        MoViNet \cite{kondratyuk2021movinets}         & \textbf{47.7}  & \textbf{72.2}  & 57.3  &          \\ 
        \hline
        ViViT-L          & 44    & 66.4  & 56.8  & K400     \\ 
        \hline
        TSformer         & 38.05 & 60.21 & 42.36 & SSv2     \\ 
        ours-TSformer    & \textbf{40.7}  & \textbf{62.38} & \textbf{45.84} & SSv2     \\ 
        \hline
        Swin-S           & 43.1  & 64.32 & 57.14 & K400     \\ 
        Ours-Swin-S      & \textbf{44.02} & \textbf{65.86} & \textbf{58.4}  & K400     \\ 
        \hline
        Swin-B           & 44.7  & 68.47 & 58.6  & K400     \\ 
        Ours-Swin-B      & \textbf{44.9}  & \textbf{69.3}  & \textbf{58.7}  & K400     \\ 
        \hline
        Mformer-HR       & 44.5  & 67    & 58.5  & K400     \\ 
        ORViT Mformer-HR & 45.7  & 68.4  & 58.7  & K400     \\ 
        Ours-Mformer-HR  & \textbf{46.3} & \textbf{69.85} & \textbf{59.07} & K400     \\ 
        \hline
\end{tabular}
\end{center}
\end{table}

\begin{table}
\begin{center}
\caption{\textbf{EAR results on the EGTEA Gaze+.} First block presents the results of typical CNN-based models, while other blocks present the comparison results of previous transformers and our EgoViT-based transformers.}
\label{tab:egtea}
\begin{tabular}{|l|c|c|c|}
        \hline
        Models          & top1 acc. & top5 acc. & mean acc. \\
        \hline
        Ego-RNN \cite{sudhakaran2018attention}        & 60.8     &    -      &   -     \\
        LSTA \cite{sudhakaran2019lsta}         & 61.9     &    -      &   -     \\
        Slowfast \cite{feichtenhofer2019slowfast}     & 49.16    & 71.27    & 37.3 \\
        SAP \cite{wang2020symbiotic}             & 62.7     &    -      &   -     \\
        min et al.\cite{min2021integrating}      & 69.58    &    -      & 62.84 \\
        \hline
        TSformer        & 62.61    & 91.79    & 54.81 \\
        ours-TSformer   & \textbf{63.71}    & \textbf{91.8}     & \textbf{56.8} \\
        \hline
        Mformer-HR      & 65.84    & 92       & 57.9 \\
        Ours-Mformer-HR & \textbf{66.3}     & \textbf{92.3}     & \textbf{58.92} \\
        \hline
        Swin-S          & 67.31    & 92.88    & 60.95 \\
        Ours-Swin-S     & \textbf{67.36}    & \textbf{93.62}    & \textbf{61} \\
        \hline
        Swin-B          & 69.25    & 95.06    & 62.38 \\
        Ours-Swin-B     & \textbf{69.72}    & \textbf{96.31}    & \textbf{63.04} \\
        \hline
\end{tabular}
\end{center}
\end{table}

Tab.~\ref{tab:model_size} lists the pretrained models, which were used to initialize the model weights, together with the number of views during inference time, which was kept the same as the setting of the corresponding transformer.~In Tab.~\ref{tab:model_size}, we also compare the number of training parameters and GFLOPS of each model. It can be seen that our model always has less GFLOPS, compared to the corresponding baseline,  which helps speeding up the training significantly, and also saves from hardware memory.~Also, we can reduce the number of parameters for Swin-S and Swin-B, and incur only a small increase (9\% and 1.6\%) compared to the TimeSformer and Mformer. ORViT is another transformer, which also uses additional detection information as the input source for action recognition. Although it can help improve the performance of Mformer, it increases the number of parameters and GFLOPs significantly. Compared to  ORViT, our model is much smaller and generates better results at the same time.

\begin{table}
\begin{center}
\caption{Transformer model size comparison and the number of views during inference time.}
\label{tab:model_size}
\begin{tabular}{|l|c|c|c|}
        \hline
        Models           & \#params & GFLOPs & views \\
        \hline
        ViViT-L          & 311M     & 3992   & 4x3   \\
        \hline
        TSformer         & \textbf{121M}     & 1581   & 3x1   \\
        Ours-TSformer    & 132M     & \textbf{885}    & 3x1   \\
        \hline
        Swin-S           & 53M      & 166    & 4x3   \\
        Ours-Swin-S      & \textbf{50M}      & \textbf{93}     & 4x3   \\
        \hline
        Swin-B           & 104M     & 282    & 4X3   \\
        Ours-Swin-B      & \textbf{88M}      & \textbf{159}    & 4X3   \\
        \hline
        Mformer-HR       & \textbf{119M}     & 958.8  & 3x1   \\
        ORViT Mformer-HR & 148M     & 1259   & 3x1   \\
        Ours-Mformer-HR  & 121M     & \textbf{775}    & 3x1   \\
        \hline
\end{tabular}
\end{center}
\end{table}

\subsection{Qualitative Results}
\label{ssec:Qualitative}
\begin{figure*}
  \centering
    \includegraphics[width=1\linewidth]{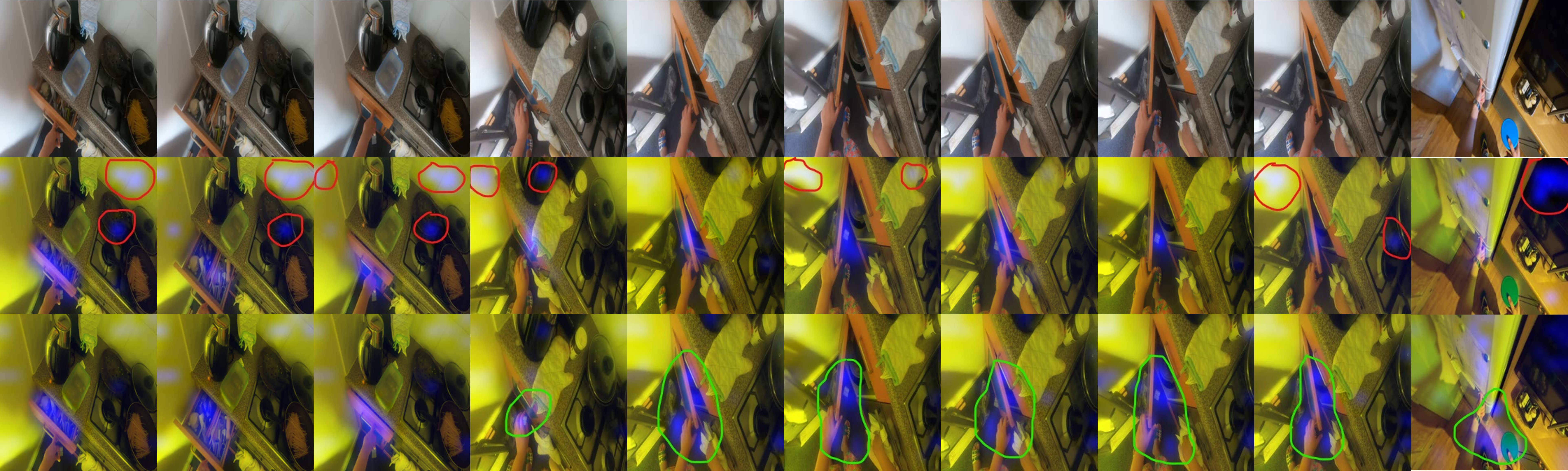}
  \caption{\textbf{Visualization of spatial attention distribution.} Top to bottom row: RGB frames from the original video, the frames with the attention map generated by TimeSformer and TimeSformer equipped with the proposed DCTG, respectively.}
  \label{fig:DCTG_attention2}
\end{figure*}

To provide evidence that our proposed DCTG can truly help the model focus on hand-objects interactions when inferring an egocentric action, we compare the spatial attention weights of TimeSformer with and without the DCTG, i.e. we compare the static class token versus the dynamic class token. We extract and visualize the attention map, which is calculated when the CT is treated as the query in the last self-attention block of the model, by normalizing it along the spatial axes. This way, we can see the weight contribution of the informative token at each location to the class token. The results are shown in Fig.~\ref{fig:DCTG_attention}. As can be seen, with our DCTG module, the model tends to focus more on the action-related parts, i.e. the hand-object interactions (marked with green ellipses), and gives less attention to the insignificant parts (marked with red ellipses). More examples are provided in Fig.~\ref{fig:DCTG_attention2}.

\begin{figure*}
    \centering
    \begin{minipage}[b]{1.0\linewidth}
        \begin{minipage}[b]{1.0\linewidth} \centering \textbf{\footnotesize{Take eating utensil.} }\end{minipage}
    \end{minipage}
    \begin{minipage}[b]{1.0\linewidth}
        \includegraphics[width=1.0\linewidth]{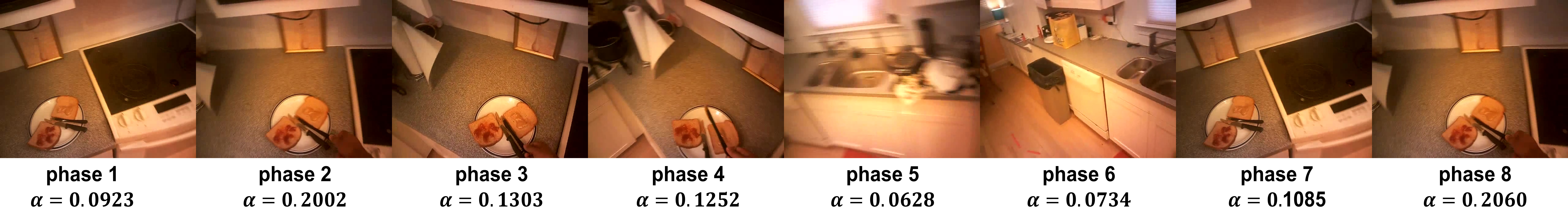}
    \end{minipage}
    \begin{minipage}[b]{1.0\linewidth}
        \begin{minipage}[b]{1.0\linewidth} \centering \textbf{\footnotesize{Cut tomato.} }\end{minipage}
    \end{minipage}
    \begin{minipage}[b]{1.0\linewidth}
        \includegraphics[width=1.0\linewidth]{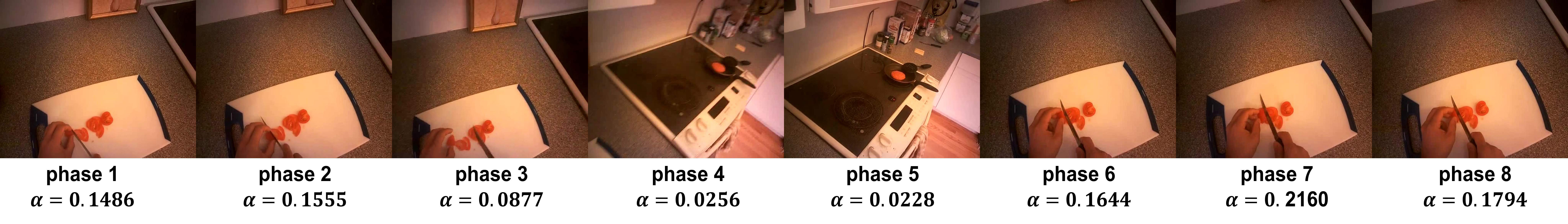}
    \end{minipage}
    \begin{minipage}[b]{1.0\linewidth}
        \begin{minipage}[b]{1.0\linewidth} \centering \textbf{\footnotesize{Turn on faucet.} }\end{minipage}
    \end{minipage}
    \begin{minipage}[b]{1.0\linewidth}
        \includegraphics[width=1.0\linewidth]{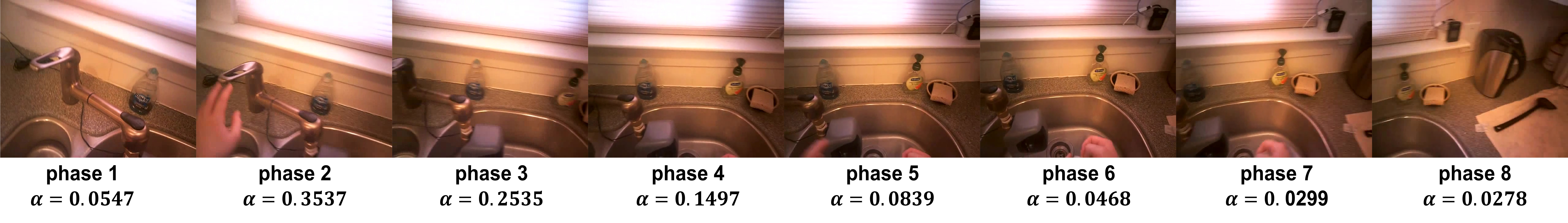}
    \end{minipage}
    \begin{minipage}[b]{1.0\linewidth}
        \begin{minipage}[b]{1.0\linewidth} \centering \textbf{\footnotesize{Read recipe.} }\end{minipage}
    \end{minipage}
    \begin{minipage}[b]{1.0\linewidth}
        \includegraphics[width=1.0\linewidth]{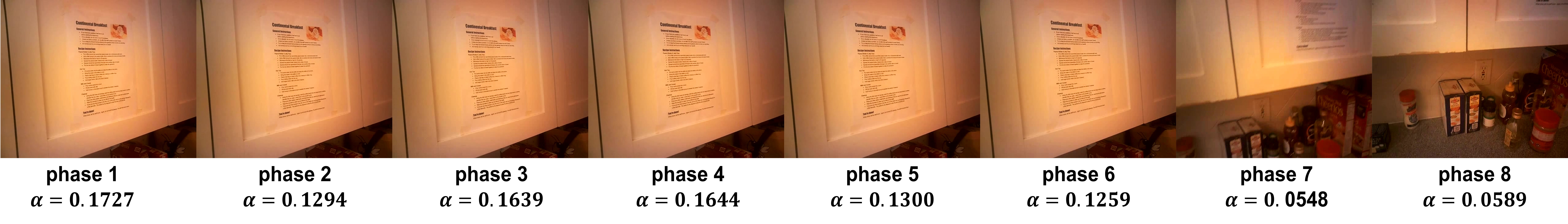}
    \end{minipage}
    \caption{Visualization of phase attention distribution.}
    \label{fig:PADM_Attention2}
\end{figure*}

To illustrate that the dynamic merging module can properly tune the attention on class tokens of different phases, we show the calculated score for each phase of various inputs. The results in Fig.~\ref{fig:phase_attn} are produced by the TimeSformer with only PADM architecture (G=8, DR=2). It can be seen that the mechanism can always generate higher scores for the important phases ($2^{nd}$, $3^{rd}$, $4^{th}$, and $5^{th}$ phases of the ``take pasta-container" action; $5^{th}$, $6^{th}$, $7^{th}$, and $8^{th}$ phases for the ``close fridge" action). More examples are provided in Fig.~\ref{fig:PADM_Attention2}. It can be seen that the mechanism can always generate higher scores for the important phases. For instance, phases 2, 3 and 8 for ``Take eating utensil", phase 2 of ``Turn on faucet" and phases 1, 2, 6, 7 and 8 of ``Cut tomato" have high scores, and correspond to the important parts of the video most relevant to the corresponding action. 
\begin{figure}[t!]
  \centering
  \begin{minipage}[b]{1.0\linewidth}
  \begin{minipage}[b]{0.24\linewidth} \centering \textbf{\footnotesize{Swin} }\end{minipage}
  \begin{minipage}[b]{0.24\linewidth} \centering \textbf{\footnotesize{Swin} }\end{minipage}
  \begin{minipage}[b]{0.24\linewidth} \centering \textbf{\footnotesize{TimeSformer}} \end{minipage}
  \begin{minipage}[b]{0.24\linewidth} \centering \textbf{\footnotesize{TimeSformer}} \end{minipage}
  \end{minipage}
\begin{minipage}[b]{1.0\linewidth}
  \includegraphics[width=0.24\linewidth]{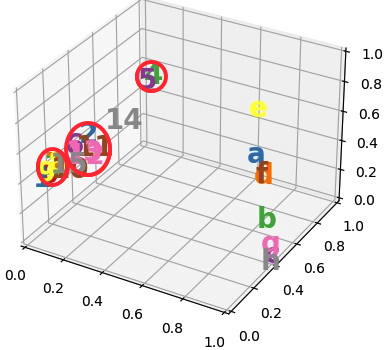}
  \includegraphics[width=0.24\linewidth]{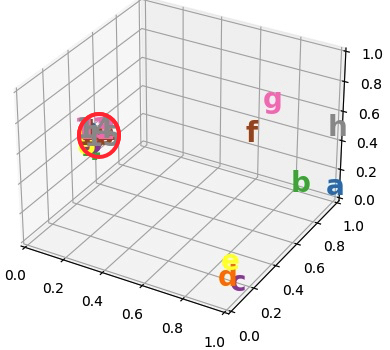}
  \includegraphics[width=0.24\linewidth]{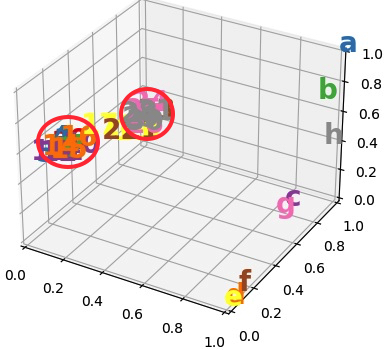}
  \includegraphics[width=0.24\linewidth]{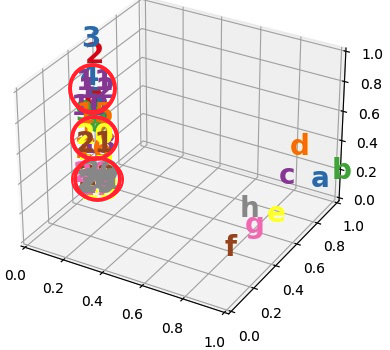}
  \end{minipage}
  \caption{\textbf{Visualization of temporal vectors.} We use the numbers and letters to indicate the temporal ID of the feature vectors for baselines and our EgoViT, respectively. Letters ``a" to ``h" represent the $1^{st}$ to $8^{th}$ temporal positions, respectively.}
  \label{fig:t_vec}
\end{figure}

To prove that there is still redundant information along the temporal axis with the previous video transformers, we visualize the distribution of average tokens at each temporal position, in the feature space of the last block, by using PCA dimension reduction algorithm. We compare our EgoViT with TimeSformer and Swin. There are 32, 16, and 8 temporal positions at the last block of TimeSformer, Swin, and our EgoViT, respectively. As shown in Fig.~\ref{fig:t_vec}, there are many overlapping points in TimeSformer and Swin. Most of them occur in consecutive frames, indicating that consecutive frames carry similar semantic meaning, and previous models cannot filter the redundant information out. For our EgoViT, there is less overlap in the feature space, which means that although its temporal dimension is smaller, it can retain complementary features and still capture rich semantic information.

%% file: 5_conclusion.tex
We have proposed a new method, referred to as EgoViT, which can be incorporated with different video transformers for egocentric action recognition. We have introduced a Dynamic Class Token Generator (DCTG) that leverages the pre-extracted hand-object interaction features to dynamically generate a class token for each video. We have shown that the DCTG is more effective than the static class token used in previous transformers. We have also presented a pyramid architecture with dynamic merging module, which can properly model the temporal relationship and reduce the redundant information that the traditional video transformer do not filter out. We have demonstrated the effectiveness and efficiency of our EgoViT by comparing it with the most well-known video transformers both quantitatively and qualitatively. Our proposed DCTG can be seen as a fusion scheme for video transformers. Therefore, an interesting extension is to inject DCTG with other kinds of information, such as optical flow and audio features, which will be pursued in our future work. In addition, although our EgoViT can boost the performance of all the video transformers tested, and outperform the current best CNN model, MoViNet, in noun prediction on the EK100 dataset, it cannot yet surpass MoViNet for action and verb class prediction. In our future work, we will improve our model further by investigating transformers and injecting DCTG with other information, as mentioned above.